\documentclass[conference]{IEEEtran}

\usepackage{cite}

\ifCLASSINFOpdf
   \usepackage[pdftex]{graphicx}
   \usepackage[space]{grffile}
\else
\fi
\usepackage{amsmath}
\usepackage{textcomp}

\begin{document}

\title{WiFi based trajectory alignment, calibration and crowdsourced site survey using smart phones and foot-mounted IMUs}

\author{\IEEEauthorblockN{Yang Gu\textsuperscript{1,2}, Caifa Zhou\textsuperscript{1}, Andreas Wieser\textsuperscript{1}}
\IEEEauthorblockA{Institute of Geodesy and Photogrammetry\textsuperscript{1}\\
ETH Zurich\\
Zurich, Switzerland\\
\{yang.gu, caifa.zhou, andreas.wieser\}@geod.baug.ethz.ch}
\and
\IEEEauthorblockN{Zhimin Zhou\textsuperscript{2}}
\IEEEauthorblockA{College of Electronic Science and Engineering\textsuperscript{2}\\
National University of Defense Technology\\
Changsha, Hunan, China\\
zhouzhimin@nudt.edu.cn}}
\maketitle

\begin{abstract}
Foot-mounted inertial positioning (FMIP) can face problems of  inertial drifts and unknown initial states in real applications, which renders the estimated trajectories inaccurate and not obtained in a well defined coordinate system for matching trajectories of different users.  In this paper, an approach adopting received signal strength (RSS) measurements for Wifi access points (APs) are proposed to align and calibrate the trajectories estimated from foot mounted inertial measurement units (IMUs). A crowdsourced radio map (RM) can be built subsequently and can be used for fingerprinting based Wifi indoor positioning (FWIP). The foundation of the proposed approach is graph-based simultaneously localization and mapping (SLAM). The nodes in the graph denote  users' poses and the edges denote the pairwise constrains between the nodes. The constrains are derived from: (1) inertial estimated trajectories; (2) vicinity in the RSS space. With these constrains, an error functions is defined. By minimizing the error function, the graph is optimized and the aligned/calibrated trajectories along with the RM are acquired. The experimental results have corroborated the effectiveness of the approach for trajectory alignment, calibration as well as RM construction.
\end{abstract}

Keywords-\textit{indoor positioning; trajectory alignment/calibration; radio map construction; graph based SLAM}

\IEEEpeerreviewmaketitle

\section{Introduction}
The combination of foot-mounted inertial positioning (FMIP) and fingerprinting based Wifi indoor positioning (FWIP) can have some interesting use cases where none of the single perspective positioning solutions are suitable. For example, at airports or malls, there is a need for both personnel tracking (e.g. security staff) and providing an indoor positioning service for passengers and customers. The latter is preferably achieved by FWIP using the persons' own Wifi enabled mobile devices while requires construction and update of a radio map (RM).

FMIP can be used for pedestrian tracking with an inertial measurement unit (IMU) mounted on the foot. FMIP is essentially inertial  based pedestrian dead reckoning (PDR), and it has two inherent challenges: (1) accumulating positioning errors, (2) unknown initial state. The former challenge is a severe limitation, and normally additional observations are needed for PDR. A problem called trajectory alignment arises from the second challenge, which is necessary to solve in order to obtain a user's trajectory in a well defined coordinate system and thus allow offering location based services (LBS) or matching trajectories of different users.

FWIP is one of the most promising solutions for indoor positioning. One of the reasons is that Wifi enabled mobile devices and Wifi signals are almost ubiquitous available in indoor environments. In a Wifi based RM, the received signal strength (RSS) from different access points (APs) are associated with positions in a coordinate system. The associated RSS and positions are called fingerprints. With a proper RM the user can be located by matching the newly collected RSSs with those stored in the RM adopting algorithms like k-nearest neighbour (kNN) \cite{6047914}. However, establishing the RM prior to positioning (site survey) is normally slow and labor-intensive\cite{7174948}.  The efficiency for site survey can be boosted by adopting crowdsourced approaches using data collected by the users' mobile devices\cite{6216368}\cite{Park:2010:GOI:1814433.1814461}.  Rather than requiring common users to upload and share their data, we propose to use the data collected by special users for RM creation and update, specifically of users like security staff, equipped with FMIP and moving within the indoor environment anyway. 

In this paper, we propose an approach for trajectory calibration and alignment for multiple agents using Wifi fingerprints. In our use case, the initial states (including initial positions and headings) for different agents and the RM are unknown. This approach can align the trajectories into the same coordinate system and can suppress positioning error growth due to inertial drifts. A Wifi based RM is built by adopting the crowdsourced trajectories.  The approach is under the framework of graph based simultaneously localization and mapping (SLAM), where the  vicinity derived from measured fingerprints in the RSS space and motion information derived from foot-mounted IMUs are used as constrains in optimizing the graph. Experimental results have been used to validate the effectiveness of the approach for trajectory alignment, calibration and site survey.

The structure for the remaining paper is as follows. Section 
\uppercase\expandafter{\romannumeral2} gives an overview for the two types of Wifi based SLAM approaches. Section 
\uppercase\expandafter{\romannumeral3} is the detailed description of the proposed approach. Section 
\uppercase\expandafter{\romannumeral4} is the experiments for the proposed approach and section 
\uppercase\expandafter{\romannumeral5} is the conclusion.

\section{Background}
The group of algorithms for Wifi based SLAM\cite{Durrant-whyte_1simultaneouslocalisation}\cite{bailey2006simultaneous} can be classified into two categories:  recursive SLAM and full SLAM. Normally, recursive SLAM assumes a first-order Markov model, i.e. the user's current position is only dependent on the previous one.  By recursively estimating (in a "predict-update" manner) the posterior distribution of the user's pose using a Kalman-type filter or a particle filter, this type of methods is suitable for on-line implementation. Full SLAM makes use of all past observations and  is normally solved as an optimization problem which minimizes a pre-defined overall error function. It is thus applied off-line.

\subsection{Wifi based Recursive SLAM}
In \cite{faragher2013smartslam}, the SmartSLAM method is proposed, which exploits a new
intelligent filtering approach to lower the computational cost for on-line processing. It sets a couple of rules to make the filter automatically change between an extended Kalman filter (EKF) and a distributed particle filter (DPF). In \cite{7346951}, Gaussian processes (GPs) are adopted to model the RSS observations, which can be used to calculate likelihoods for each particle in the particle filter. Another particle filter based method is proposed in \cite{Gu2017} reducing the high computational cost by GPs. The different filtering techniques used in these publications share the same goal for (1) approximating the accuracy of the full SLAM methods, and (2) lowering computational cost for on-line processing. However, they are not suitable for building crowdsourced RM, because pose relationships between different trajectories are unknown and thus there can be no prediction step in the filtering process. 

\subsection{Wifi based full SLAM}
Graph based SLAM\cite{5681215} is a classical realization for the full SLAM problem. The nodes in the graph denote poses and the edges denote constraints between the poses derived from measurements. The optimization of the graphs is essentially finding the configuration of nodes which satisfies the measurements best. Researchers in \cite{6817853} modified the graph based SLAM by creating "virtual" nodes as landmarks defined by average positions of poses whose observations are similar in RSS space. In \cite{Nowicki2015}, the graph is constructed by adding nodes whose RSS observations and locations are known a priori. However, these existing full SLAM methods are not designed for multiple trajectories, especially when the initial poses for these trajectories are different and unknown. Although in \cite{7277082}, the researchers  constructed crowdsourced RMs, the trajectories are pre-aligned using a trusted portable navigator (TPN), which may not be available in many applications.

The proposed approach can align and calibrate trajectories from multiple agents, and it can build a crowdsourced RM subsequently without  need for initial poses and ground truth reference positions.

\section{Method}
\subsection{Fundamentals}
Given an initial position and heading, FMIP can provide a PDR trajectory. Normally, the estimated position and heading are updated each time a new step is detected by the foot-mounted IMU. A state vector containing the horizontal positions and the heading is defined as $\mathbf{pos}_{k}^{i}=(x_{k}^{i},y_{k}^{i},\theta _{k}^{i})$, where the superscript  denotes the $i^{\text{th}}$ trajectory  and the subscript denotes the time index. Then the  trajectory can be represented as:
\begin{equation}
\mathbf{TRAJ}^{i}=(\mathbf{pos}_{1}^{i},\mathbf{pos}_{2}^{i},...,\mathbf{pos}_{k}^{i},...)\
\end{equation} 
For different trajectories, as the actual initial state can be different and unknown, a default initial state is set as: 
\begin{equation}
\mathbf{pos}_{1}^{i}=(0,0,\psi)
\end{equation}
$\psi$ is a random variable which represents initial heading of the user. This is due to two reasons: (1) we set the IMU's initial heading as 0 during inertial calculation; (2) the difference between IMU's initial heading and user's initial heading is unknown. In this configuration, the trajectories are not in the same coordinate system and the coordinate transformation parameters are initially unknown. 

Since pose and RSS observations are acquired at different rates, a scheme in \cite{Gu2017} is adopted to associate each $\mathbf{pos}_k^i$ with a set of RSS observations $\mathbf{RSS}_k^i$.

The goal of graph-based SLAM can be rephrased as finding a sequence of nodes $\mathbf{pos}_{1:N}$ which minimizes the overall error function
\begin{equation} \label{eq: errorfunc}
	F=F_1+F_2	
\end{equation}
where
\begin{multline}\label{eq: F1}
	F_1=\sum_{k=1}^{N-1}[\mathbf{e}_{\mathrm{IMU}}^\mathrm{T}(\mathbf{pos}_{k-1,k},\mathbf{u}_{k-1,k})\mathbf{\Omega}^{\mathrm{IMU}}\\
	\mathbf{e}_{\mathrm{IMU}}(\mathbf{pos}_{k-1,k},\mathbf{u}_{k-1,k})]
\end{multline}
and 
\begin{equation}\label{eq: F2}
	F_2=\sum_{k=1}^N\sum_{q=1}^{N}\Omega^\mathrm{Wifi}{e}_{\mathrm{Wifi}}(\mathbf{pos}_k,\mathbf{pos}_q,\mathbf{RSS}_k,\mathbf{RSS}_q)
\end{equation}
$N$ is the total number of poses within a specific trajectory (superscript denoting different trajectories omitted for readability). The error function is comprised of two parts. The first part $F_1$ are errors derived from measurements of the foot-mounted sensor through PDR, while the second part $F_2$ are errors derived from the RSS measurements. The two types of errors also correspond to two error functions $\mathbf{e}_\text{IMU}(.)$ (vector output) and $e_\text{Wifi}(.)$ (scaler output) respectively, and the graph’s edges are classified into IMU edges and Wifi edges accordingly. Other details of the cost function will be given in the next subsection.

The problem of graph-based SLAM is essentially comprised of two steps: graph formation and graph optimization. Our main contribution lies in the first step where the nodes, edges and error functions are defined. In this paper, only a simple review is given for the graph optimization step, which we treat as a Least Squares Optimization (LSO) problem, which can be solved via iterative local linearizations using the Gauss-Newton or Levenberg-Marquardt algorithms\cite{Lourakis:2009:SSP:1486525.1486527}. In our implementation, we use the g2o graph optimization framework\cite{5979949}. In the optimization process, we set the first pose in one of the many trajectories as a reference (invariant during optimization), while other poses are variant.
\subsection{Graph Formation}
\subsubsection{Constrains derived from the foot sensor}
From the foot-mounted module, constrains between adjacent poses are formed. 
\begin{equation}\label{eq: motion model}
	\begin{pmatrix}
	x_k\\
	y_k\\
	\theta_{k}
	\end{pmatrix}
	=
	\begin{pmatrix}
	x_{k-1}\\
	y_{k-1}\\
	\theta_{k-1}
	\end{pmatrix}
	+
	\begin{pmatrix}
	\Delta x_{k-1}cos(\theta_{k-1})-\Delta y_{k-1}sin(\theta_{k-1})\\
	\Delta x_{k-1}sin(\theta_{k-1})+\Delta y_{k-1}cos(\theta_{k-1})\\	
	\Delta \theta_{k-1}
	\end{pmatrix}
\end{equation}
From the $\mathbf{pos}$ estimations provided by the foot-mounted module, a vector which represents the differences between adjacent poses can be calculated from equation (\ref{eq: motion model}).  This vector $(\Delta x_{k-1},\Delta y_{k-1},\Delta \theta_{k-1})$ corresponds to $\mathbf{u}_{k-1,k}$ in equation (\ref{eq: F1}) and it denotes the $\mathbf{pos}$ change in the IMU body frame. Then the error vector $\mathbf{e}_{\mathrm{IMU}}(.)$ in equation (\ref{eq: F1}) can be represented as:
\begin{equation}
	\mathbf{e}_{\mathrm{IMU}}(.)=\mathbf{u}_{k-1,k}^\mathrm{nodes}-\mathbf{u}_{k-1,k}
\end{equation}
where the superscript \textit{nodes} denotes that the variables are derived from the pose variables $\mathbf{pos}_{k-1}$ and $\mathbf{pos}_{k}$ according to equation (\ref{eq: motion model}), while the term without the superscript denotes the measurements  from the foot-mounted sensor. 

The information matrix $\mathbf{\Omega}^\mathrm{IMU}$ models the uncertainty in $\mathbf{e}^\mathrm{IMU}$, which is related to the quality of the estimated poses from the foot-mounted module. Normally the distance related measurements are more accurate than the heading related ones, with $\mu=20$. 
\begin{equation}
	\mathbf{\Omega}^\mathrm{IMU}=
	\begin{bmatrix}
	1/m^2 & 0 & 0\\
	0 & 1/m^2 & 0\\
	0 & 0 & \mu/rad^2
	\end{bmatrix}
\end{equation}

\subsubsection{Constrains derived from Wifi fingerprints}
In FWIP a basic assumption is that vicinity in the RSS space is with vicinity in the coordinate space. This is also used in our proposal.  In the coordinate space the Euclidean distance is naturally used to define vicinity. In the RSS space, a suitable metric needs to be chosen for calculating the distance between two Wifi fingerprints $\mathbf{RSS}_k$ and $\mathbf{RSS}_q$. Firstly the RSS are completed to the same vector length such that they both have signal strength readings from the same APs. If a signal strength reading from a specific AP does not exist in one of $\mathbf{RSS}_k$ and $\mathbf{RSS}_q$, a default reading with the value of -110dB is inserted. The underlying assumption is that missing readings are buried in noise with a level of -110dB. The distance of two RSS fingerprints is then defined as:
\begin{equation}
	d(\mathbf{RSS}_k,\mathbf{RSS}_q)=\sqrt{\frac{\sum_{j=1}^{M}({RSS}_k^{(j)}-{RSS}_k^{(j)})^2}{M}}
\end{equation}
where $M$ is the number of APs and the superscript $j$ denotes the index of the AP.

\begin{figure}
	\centering
	\includegraphics[trim=1cm 1cm 1cm 1cm, width=\linewidth]{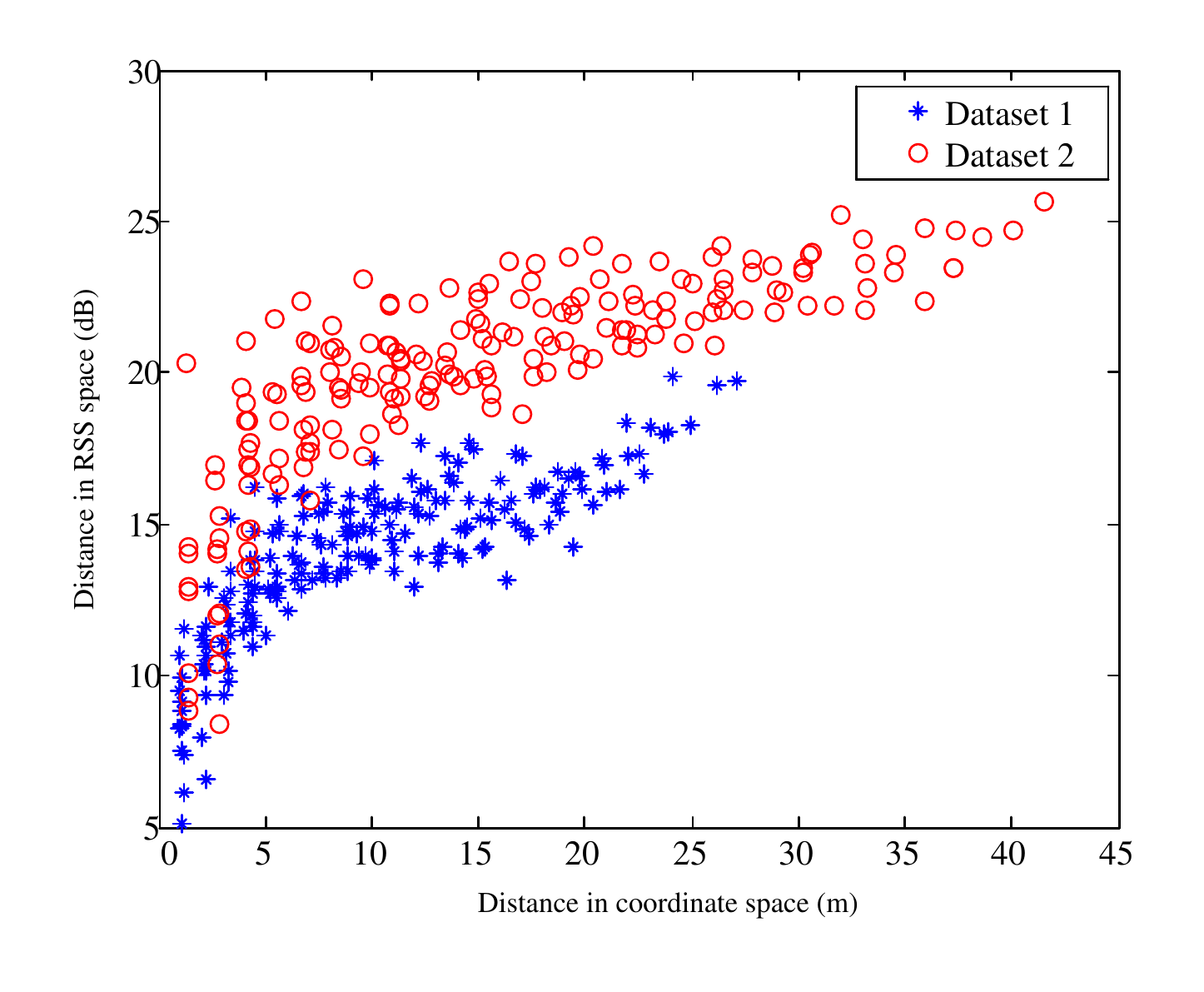}
	\caption{Distances in both coordinate space and RSS space for two datasets with different Wifi signal density. Within the datasets (both no longer than a period of several minutes), errors in position estimations are assumed to be insignificant and thus ignored.}
	\label{fig: distance comparisons}
\end{figure}

For the proposed algorithm we need to define whether two positions are close to each other or not in the RSS space. Since the relation between metric distance in coordinate space and distance in RSS space varies with APs and environment (see Fig. \ref{fig: distance comparisons}) we cannot use a fixed threshold for this decision.

Instead we determine the threshold using a sliding window over a few consecutive measurement epochs (steps). The window ranges from the $({k-L/2})^\mathrm{th}$ step to the $({k+L/2})^\mathrm{th}$ step, where $k$ is considered as the current step and $L$ is window length. All RSS measurements obtained within these steps can be regarded as readings in the same region.  We choose the median of the RSS distances within this window as the current threshold $d_{med}$. 

If the RSS distance between two fingerprints is lower than the threshold, a new edge is created between the corresponding nodes. There will of course be at least one such edge within the current window (it will later be ignored). However, the RSS distances to steps recorded long before the current window are also checked and thus edges may also be created to nodes occupied much earlier. These are the ones potentially adding important information: they indicate that the trajectory possibly passes locations covered earlier.
We then define the error functions ${e}_{\mathrm{Wifi}}(.)$ in equation (\ref{eq: F2}) which can represent the inconsistency in terms of vicinity between RSS space and coordinate space:
\begin{equation}
{e}_{\mathrm{Wifi}}(.)=
\begin{cases}
	0, &d(k,q)<d_{min}\\
	{e}_{\mathrm{Wifi}}^{max},&d(k,q)>d_{max}\\	
	{e}_{\mathrm{Wifi}}^{max}\frac{d(k,q)-d_{min}}{d_{max}-d_{min}},&\text{otherwise}
\end{cases}
\end{equation}
where $d(k,q)$ is the Euclidean distance between $\mathbf{pos}_k$ and $\mathbf{pos}_q$. For the nodes whose RSS distance are under the previously mentioned median $d_{med}$, we treat those with also very close distance in coordinate as totally consistent (error is zero) and nodes with distinctly different coordinate positions as totally inconsistent (error equals a fixed maximum). However, we also introduce some fuzziness in the vicinity of $d_{min}$ and  $d_{max}$ by linearly increasing the error. Herein $d_{min}$ is set to be 2m and $d_{max}$ is set to be 30m.

\subsubsection{Constrains between multiple trajectories}
\begin{figure}
	\centering
	\includegraphics[trim=0.5cm 1cm 0.5cm 0cm, width=\linewidth]{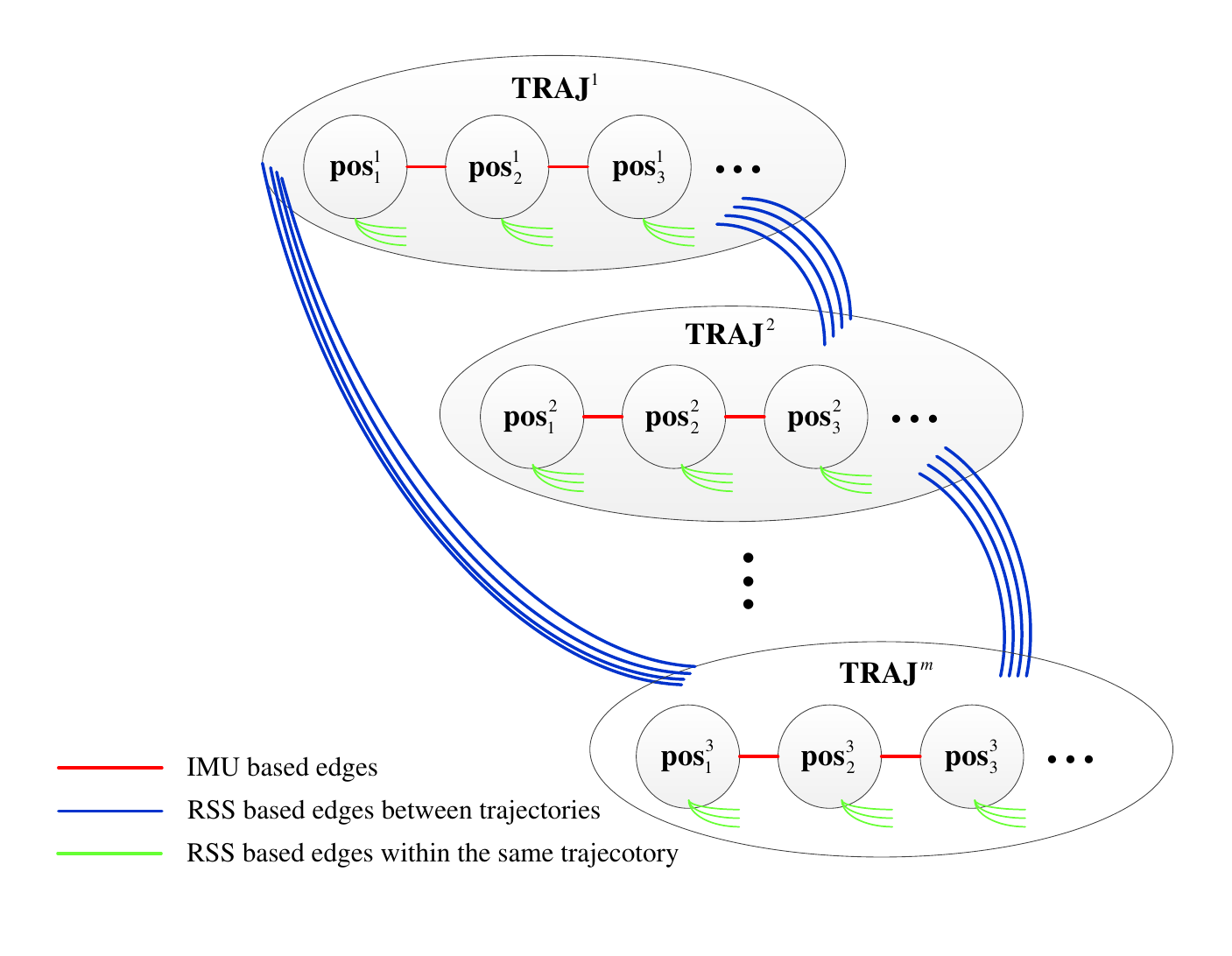}
	\caption{A illustration for graph constructed from multiple trajectories. Red lines are IMU based edges between adjacent nodes within a trajectory. Green lines are RSS based edges within a trajectory (not completely drawn). Blue lines are RSS based edges between trajectories.}
	\label{fig: multiple trajectory}
\end{figure}

In aforementioned use cases, crowdsourced data are needed for positioning multiple agents and building a radio map for a large area. As Fig. \ref{fig: multiple trajectory} shows, the graphs for multiple trajectories are built in a way which appears to be a single trajectory. The differences are twofold. (1) IMU based edges (red lines) exist between adjacent nodes from the beginning to the end within the same trajectory; they do not exist between trajectories.
(2)  We do not form Wifi based edges (green lines), if two nodes of the same trajectory are near in time or distance, because they naturally satisfy the edges as described in \cite{6088632}.   
\subsection{Systematic Framework}
\begin{figure}
	\centering
	\includegraphics[trim=0.5cm 1cm 0.5cm 0cm, width=\linewidth]{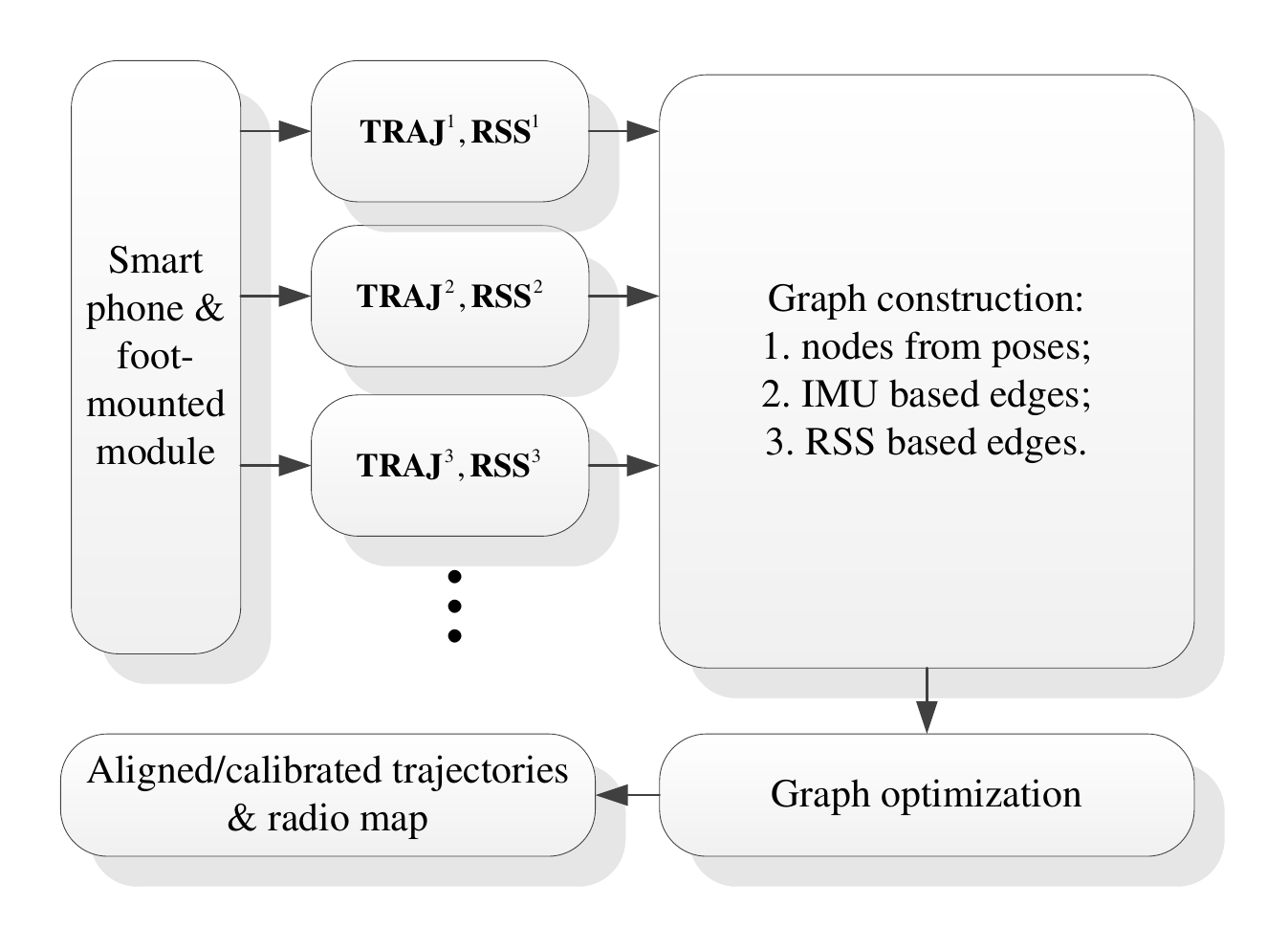}
	\caption{The systematic framework in our implementation}
	\label{fig: systematic framework}
\end{figure}

In our implementation, as shown in Fig. \ref{fig: systematic framework}, the trajectories and the RSS fingerprints are collected with a foot-mounted module and a smart phone respectively. These data are adopted to create a graph where edges are derived both from the foot module and the RSS fingerprints. By optimizing the graph, the trajectories and the RM which best explain the observations can be identified.

\section{Experiment}
Several experiments are carried out to test the performance of the proposed approach. We firstly analyse the accuracy for the aligned/calibrated trajectories and then the crowdsourced RM.
\subsection{Trajectory accuracy analysis}
\subsubsection{Alignment accuracy analysis}
\begin{figure}
	\centering
	\includegraphics[trim=0.5cm 0.7cm 0.5cm 0.5cm, width=\linewidth]{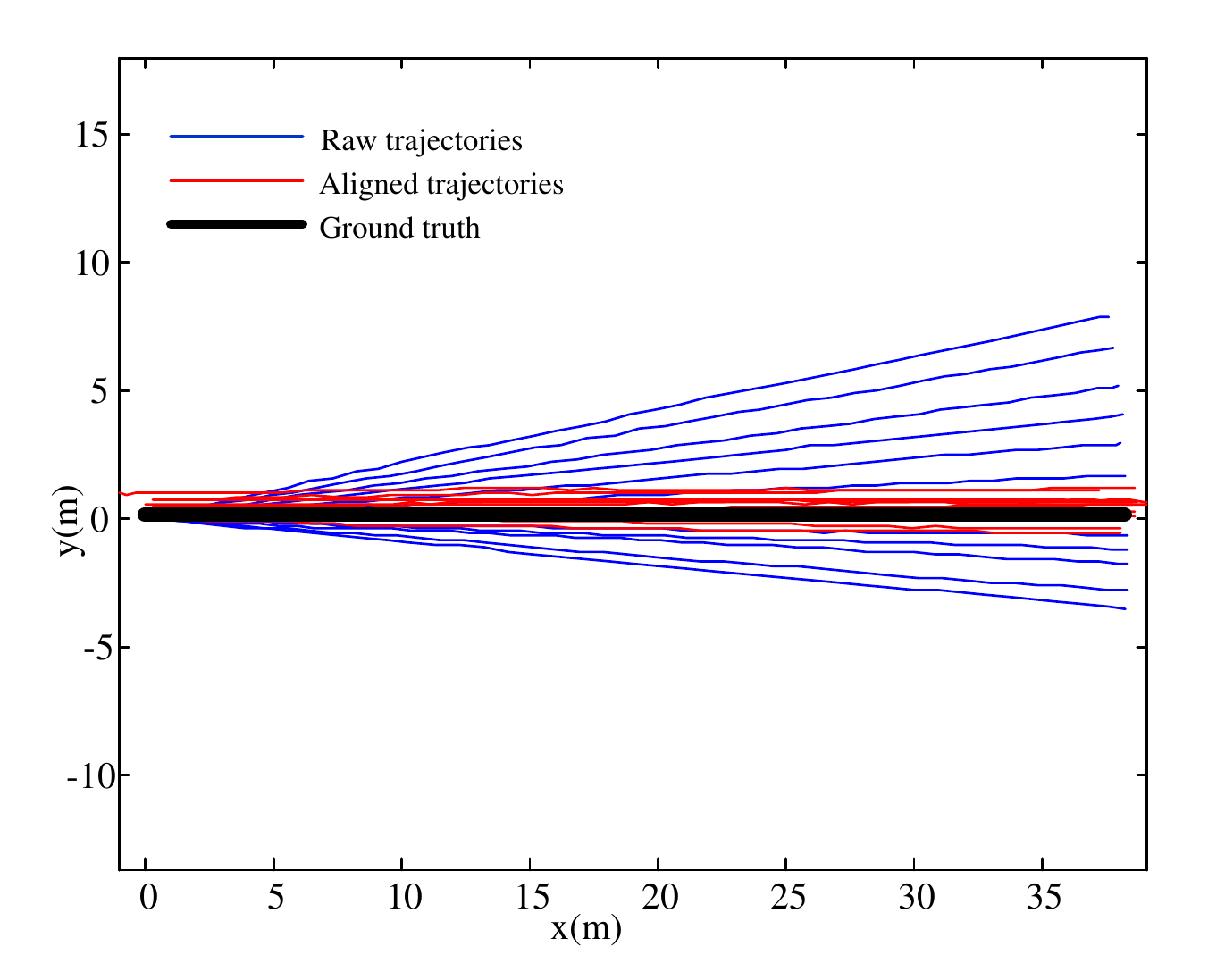}
	\caption{Alignment result for 12 straight line trajectories}
	\label{fig: alignment result}
\end{figure}

The user has walked along the same straight line trajectory (manually marked) in a corridor 12 times to test the efficiency of the proposed approach for trajectory alignment.  For clearer display, we have manually rotated one of the raw trajectories (inertial generated) to the same coordinate system as the ground truth and set the first pose of this trajectory invariant (all the other trajectories should be aligned to this one) during the graph based SLAM. As shown in Fig. \ref{fig: alignment result}, although the raw trajectories have random initial headings within a certain range, the trajectories have been successfully aligned. Compared with the ground truth, the  mean error of heading alignment is as small as about 1.0\textdegree , and the standard deviation is 0.5\textdegree.
\subsubsection{Alignment/Calibration accuracy analysis}
\begin{figure}
	\centering
	\includegraphics[trim=0.6cm 1.2cm 1cm 1cm, width=\linewidth]{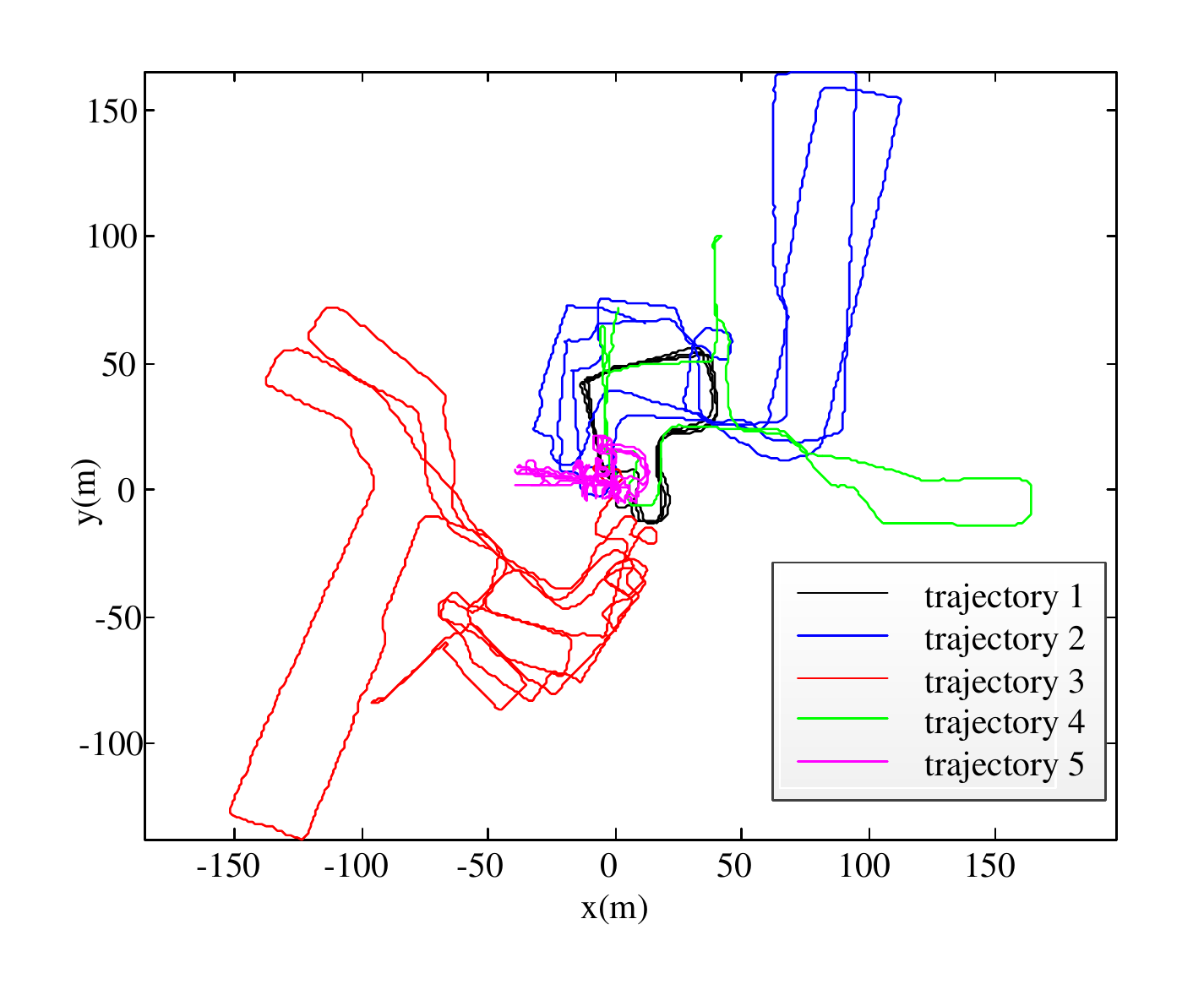}
	\caption{Five raw trajectories}
	\label{fig: multiple trajectory raw}
\end{figure}

\begin{figure}
	\centering
	\includegraphics[trim=0cm 0cm 0cm 0cm, width=\linewidth]{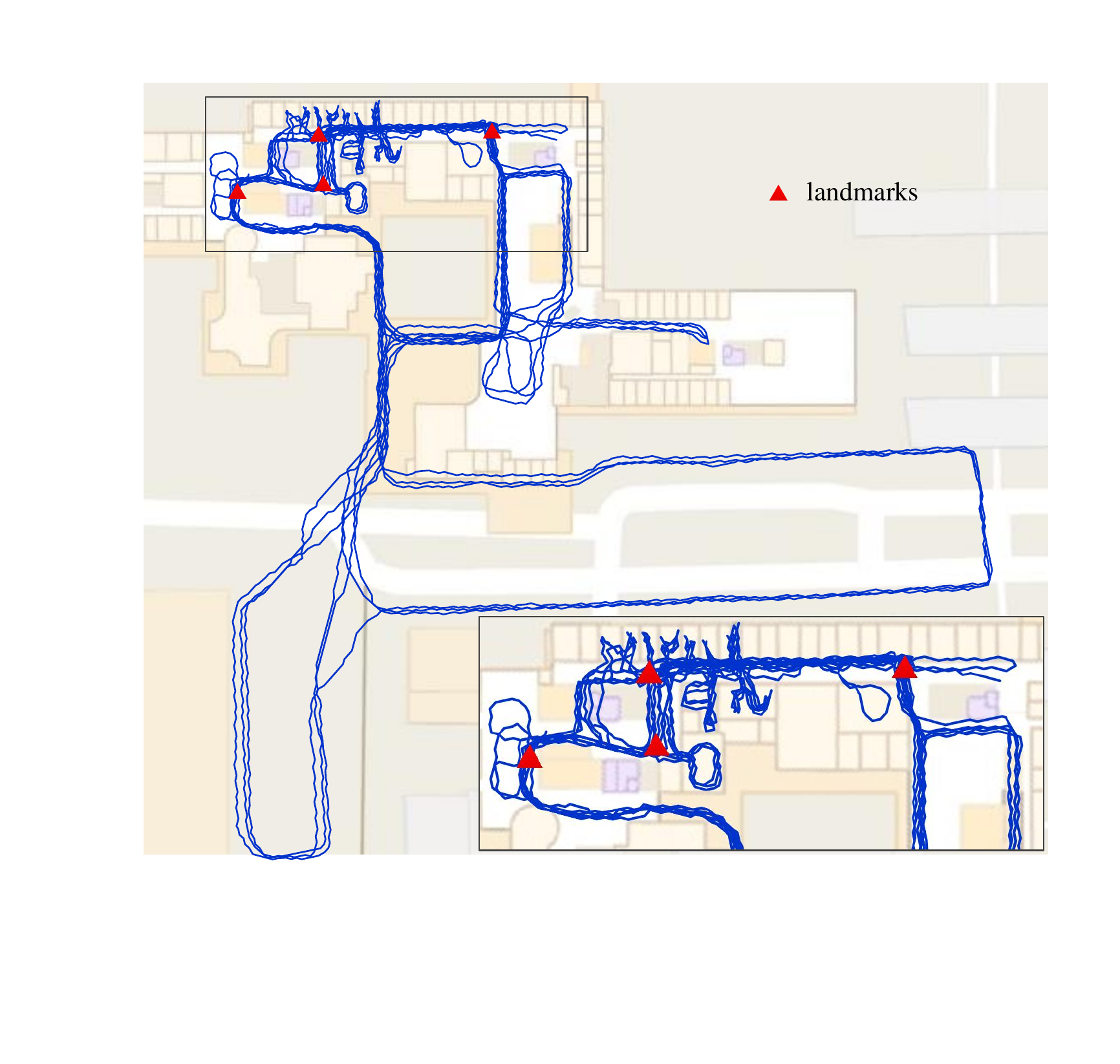}
	\caption{The aligned/calibrated trajectories over the floor plan of the experimental scene}
	\label{fig: multiple trajectory aligned}
\end{figure}

\begin{figure}
	\centering
	\includegraphics[trim=0cm 0cm 0cm 0cm, width=\linewidth]{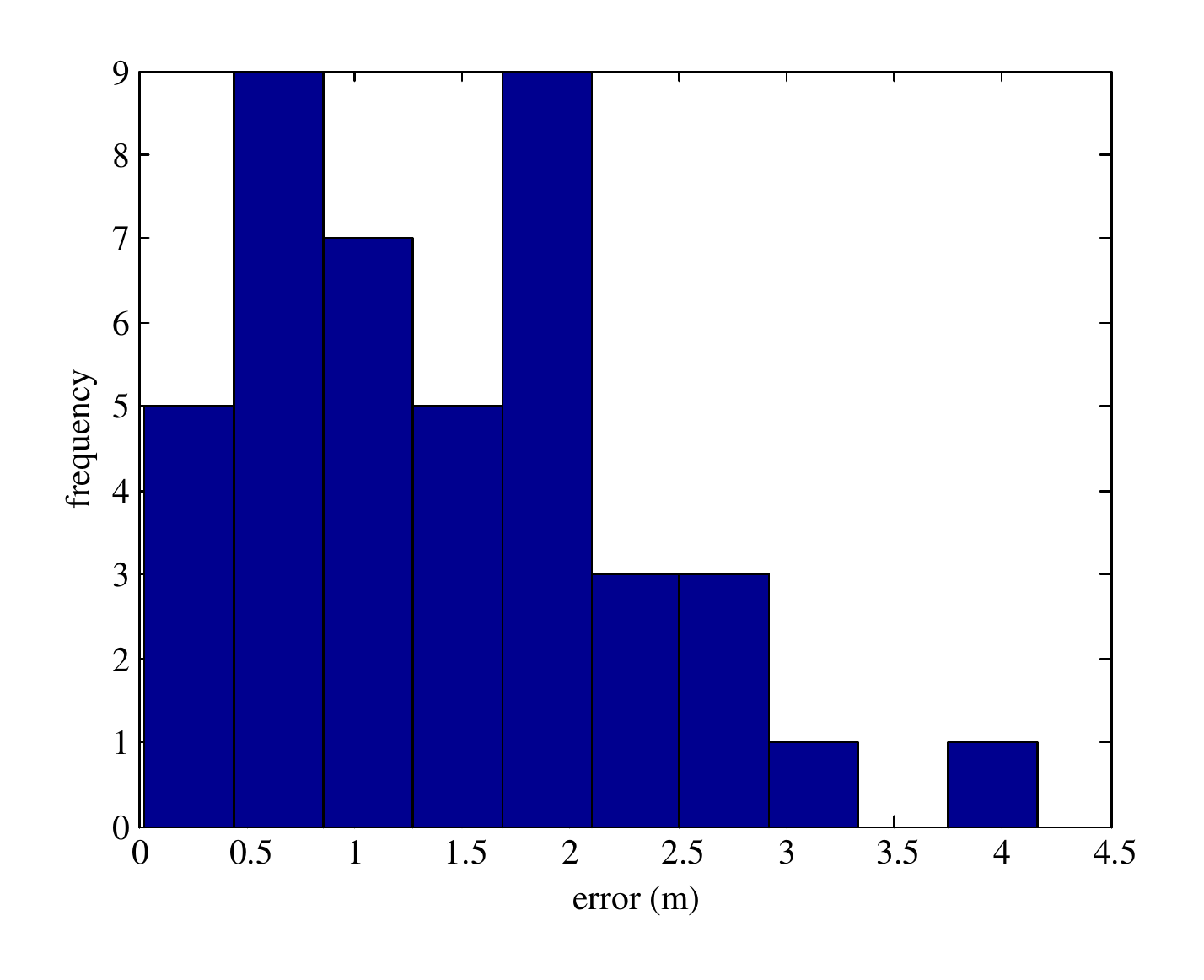}
	\caption{The histogram for positioning errors measured at the four landmarks. The errors denotes the distance between the ground truth landmarks positions and the measured positions.}
	\label{fig: errors at landmarks}
\end{figure}

To test the proposed approach, we have collected 5 long trajectories whose durations range from 9 minutes to 28 minutes with a total walking length of about 5.6 kilometers. These trajectories cover indoor and outdoor scenarios with quite different Wifi signal densities.  We have manually set 4 landmarks in the experimental scene, and their locations are measured adopting a total station. When the user walks to these landmarks, he/she is asked to press some buttons on the smart phone to record the time and landmark number. Some ground truth reference positions can be recorded consequently and positioning errors can be observed. As shown in Fig. \ref{fig: multiple trajectory raw}, the raw trajectories are not aligned and the positioning errors due to inertial drifts within some of them is obvious (e.g. the upper part of the blue trajectory should overlap).  After processing with the proposed approach, the trajectories along with the floor plan are shown in Fig. \ref{fig: multiple trajectory aligned}. Compared to the raw trajectories, the resulting ones are obtained in the same coordinate system. This can be a key requirement in applications where multiple-agent tracking is needed. Also, the processed trajectories match well with the floor plan (especially in the enlarged area where walks in different rooms are clearly separable) and the inertial drifts are significantly mitigated. The histogram for positioning errors measured at the four landmarks is shown in Fig. \ref{fig: errors at landmarks}. There are in total 43 samples collected  and the maximum error is about 4.2m. The mean error and standard deviation is 1.4m and 0.9m respectively. This further corroborates that this approach can greatly mitigate positioning errors due to inertial drift, while the positioning error reach up to approximately 30m within the raw trajectories in this experiment. 
\subsection{RM accuracy analysis}
Once the trajectory has been established, the recorded RSS values and the estimated mode positions can be used to derive a RM for fingerprinting.
An experiment is carried out to test the accuracy of this RM created from multiple trajectories (herein denoted as crowdsourced RM) using the proposed approach. We compare the crowdsourced RM to the ground truth RM, which was obtained using a smart phone and a total station for location determine with mm-accuracy. To compare the two RMs, the same test dataset of RSS values and the kNN approach is used to determine the user positions. The resulting errors were again obtained by comparisons to total station measurements.

\begin{figure}
	\centering
	\includegraphics[trim=0cm 0cm 0cm 0cm, width=\linewidth]{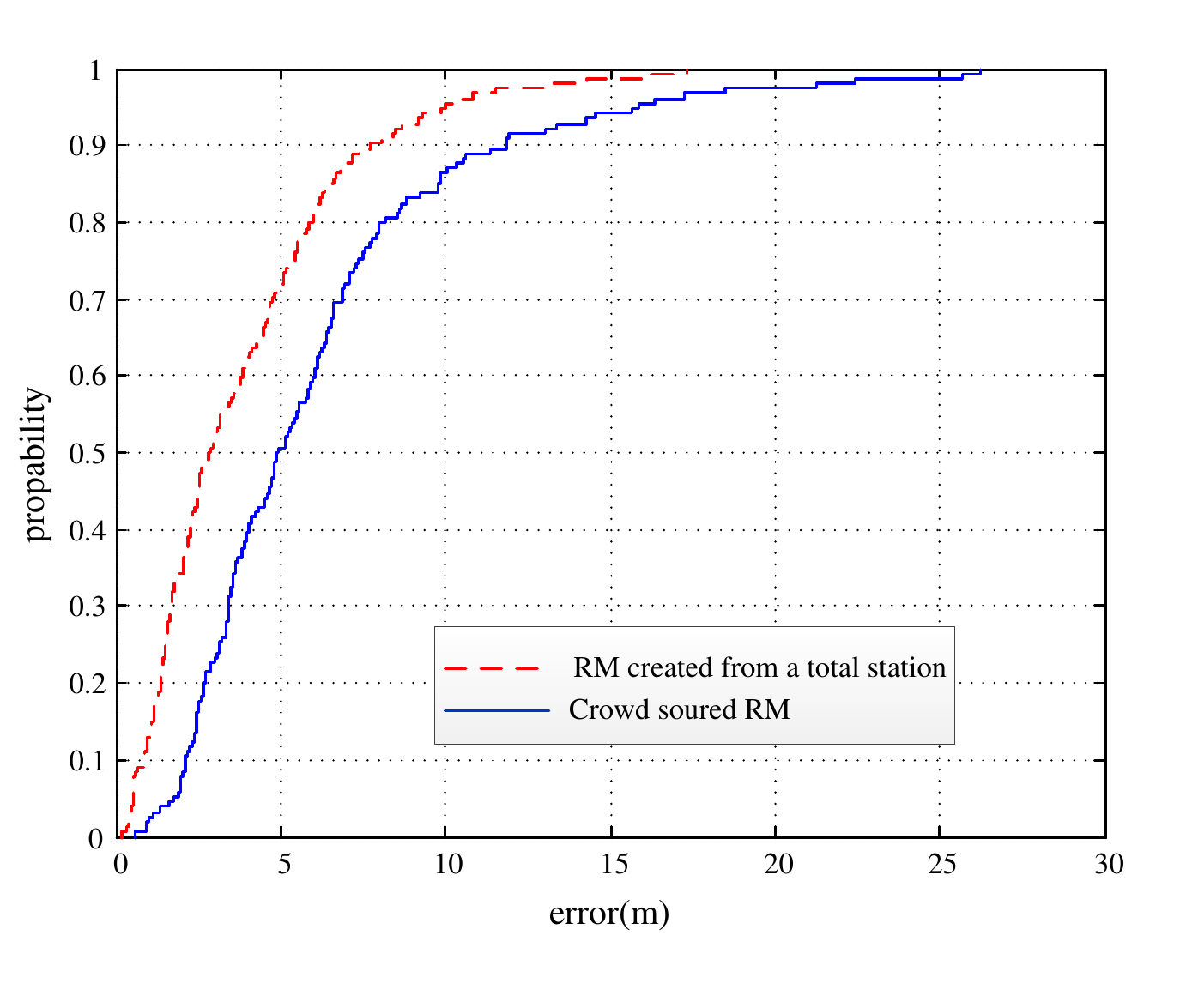}
	\caption{Cumulative distribution functions (CDFs) for the RM constructed from a total station and from the crowd sourced RM using the kNN approach.}
	\label{fig: cdfcmp}
\end{figure}

The CDFs of the positioning errors representing the quality of the RMs are shown in Fig. \ref{fig: cdfcmp}. Not surprisingly, the Wifi based positioning accuracy deteriorates when the crowdscoured RM is used; the probability of errors less than 10m drops from 0.95 to about 0.87. However, considering the scalability and convenience for constructing the RM using the proposed approach over using a total station or other dedicated measurement equipment and an off-line mapping phone, the results are very promising. Indeed the significant reduction of effort may compensate for the slight loss of accuracy. Furthermore, the accuracy can continuously  be improved by incorporating further crowdsourced datas and thus also keeping the RM up-to-date.

\section{Conclusion}
An approach is proposed for aligning and calibrating inertial generated trajectories using Wifi based RSS measurements in this paper. Additionally, the approach can also conduct easy site survey: building a RM from the crowdsourced data. Both the accuracy of the trajectories and the RM are shown in the experiment. In the multiple-trajectory alignment/calibration test, 5 trajectories with a total walking length of 5.6 kilometers are aligned/calibrated, and the mean positioning error at the pre-defined landmarks is 1.4m. Also, the crowedsourced RM is compared with the ground truth RM using the same test dataset of RSS values and the kNN approach.  The positioning error under 10m drops slightly from 0.95 to 0.87. However, considering the scalability and convenience for constructing RM using the proposed approach, the slight loss of accuracy can be compensated by the significant reduction of effort for site survey using a total station or other dedicated measurement equipment and an off-line mapping phone.


\section*{Acknowledgment}

The Chinese Scholarship Council has supported Y.G.
during his academic visit at ETH and C.Z. during his Ph.D.
studies at ETH.

\bibliographystyle{IEEEtran}
\bibliography{ipin2017}

\end{document}